\documentclass[]{mva_style}

\usepackage{graphicx}
\usepackage[sort,space,adjust,nocompress]{cite}

\usepackage[breaklinks=true,colorlinks,bookmarks=false]{hyperref}

\usepackage{amsmath}
\usepackage{amssymb}

\usepackage{multirow}
\usepackage{multicol}
\usepackage{diagbox}
\usepackage{xfrac}

\usepackage[labelformat=simple,subrefformat=simple]{subcaption}
\usepackage{caption}
\usepackage{algorithm}
\usepackage{algpseudocode}

\makeatletter
\renewcommand{\ALG@name}{アルゴリズム}
\makeatother

\newcommand{\R}{\mathbb{R}}

\usepackage{comment}

\finalcopy 

\begin{document}
\title{MoExDA: Domain Adaptation\\ for Edge-based Action Recognition}

\author{
  Takuya Sugimoto, Ning Ding, Toru Tamaki\\
  Nagoya Institute of Technology, Japan\\
  {\tt t.sugimoto.192@nitech.jp, tamaki.toru@nitech.ac.jp}\\
}

\maketitle

\section*{\centering Abstract}
\textit{
    
Modern action recognition models suffer from static bias, leading to reduced generalization performance. In this paper, we propose \emph{MoExDA}, a lightweight domain adaptation between RGB and edge information using edge frames in addition to RGB frames to counter the static bias issue.
Experiments demonstrate that the proposed method effectively suppresses static bias with a lower computational cost, allowing for more robust action recognition than previous approaches.

}

\section{Introduction}
\label{Introduction}

In recent decades, the importance of action recognition has increased, especially in surveillance, human-computer interaction, and various other domains \cite{Yu_2022_human_action_recog_survey,Kong_IJCV2022_actionrecognitionsurvey,Hutchinson_IEEEAccess2021_Action_Recognition_Survey,Hara_2018CVPR_3D_ResNet}.
However, it has been quantitatively demonstrated that current action recognition models tend to depend on \emph{static bias} \cite{Chung_NeurIPS2022_HAT,Li_ECCV2018_RESOUND,Fukuzawa_MMM2025_Zero_shot,He_ECCV2016_Workshops_withouthuman};
the excessive reliance on static appearances of the scene, such as the background and objects, rather than the dynamics of human actions.
Although static information is useful for recognition \cite{Jain_CVPR2015_WhatdoObject,Wu_CVPR2016_HarnessingObject}, static bias can lead to issues of decrease in generalization performance and limitation to particular settings.

Among the works to address this, AFD\cite{Ilic_ECCV2022_RandomDot} is a promising approach; it removed the static appearance by converting each RGB frame with random dots with the use of optical flow to warp regions of moving objects.
However, it requires a significant computational cost for random dot frames and, crucially, is not applicable to models that are trained on ordinal RGB frames.

In this study, we propose using edge frames obtained by applying lightweight edge detection to RGB frames, instead of computationally demanding warped random dot frames \cite{Ilic_ECCV2022_RandomDot}.
We use a two-stream model based on ViT \cite{Dosovitskiy_ICLR2021_ViT_Vision_transformer}
that simultaneously handles RGB and edge frames by exchanging information in the middle of the network.
To handle a domain shift that occurs between RGB and edge streams,
we propose a novel \emph{Moment Exchange Domain Adaptation (MoExDA)} module, inspired by MoEx \cite{Li_CVPR2021_MoEx}, for domain adaptation between features obtained from RGB and edge frames.
The experimental results show that the proposed model succeeded in greatly reducing the effect of static bias.

\begin{figure*}[t]
    \centering
    
    \includegraphics[width=0.8\linewidth]{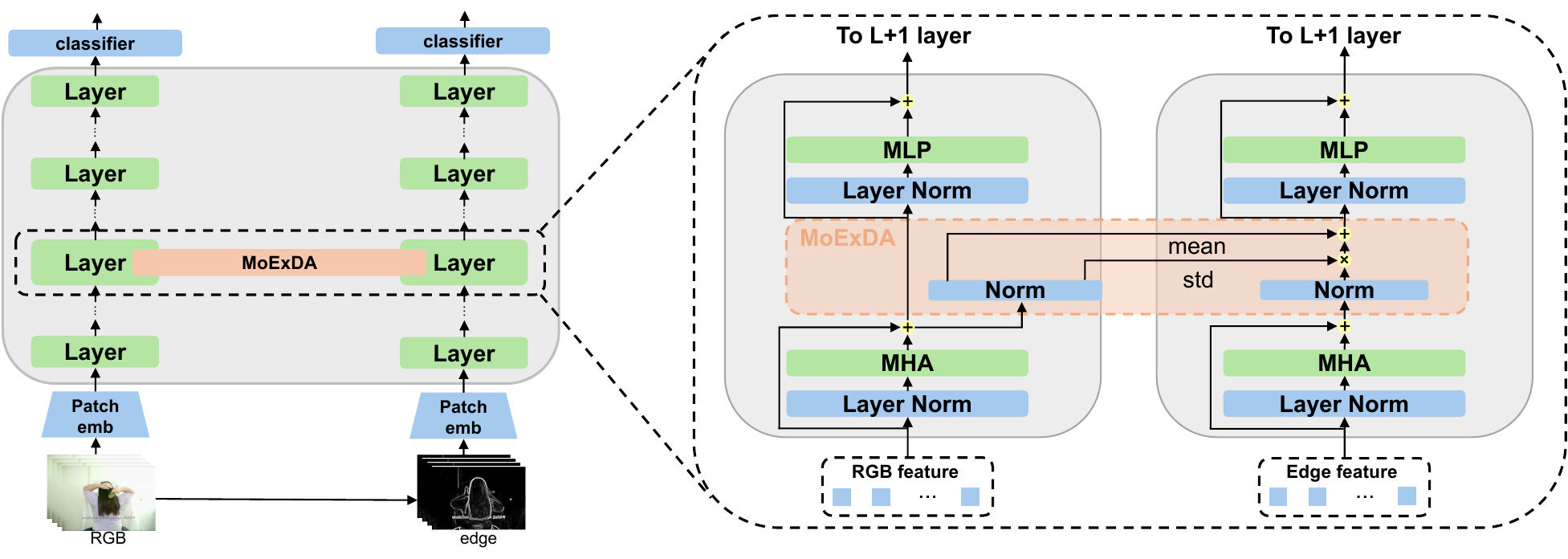}

    \caption{
        The structure of the proposed model with RGB and edge streams. The proposed MoExDA modules (orange) exchange the moments between the corresponding layers.
    }
    \label{fig:architecture_all}
    
\end{figure*}

\section{Related Work}

\subsection{Action Recognition and Static Bias}

There are various action recognition models that
take into account spatio-temporal information,
such as 3D CNN-based \cite{Carreira_2017CVPR_I3D,Feichtenhofer_2020CVPR_X3D,Feichtenhofer_2019ICCV_SlowFast,Hara_IEICE_ED2020_Action_Recognition_Survey}, ViT-based \cite{Arnab_2021_ICCV_ViVit,Bertasius_ICML2021_TimeSformer,Selva_TPAMI2023_VideoTransformerSurvey,Liu_2022CVPR_VideoSwin}, or recent CLIP-based \cite{Xu-EMNLP2021-VideoCLIP,Wang_arxiv2021_ActionCLIP,Rasheed_CVPR2023_Vifi-CLIP}.

However, it is well known that action recognition models have the tendency to rely heavily on static bias \cite{Chung_NeurIPS2022_HAT,Li_ECCV2018_RESOUND}.
The main approaches of recent work on mitigating static bias
are masking the background
\cite{Sugiura_VISAPP2024_S3Aug,Fukuzawa_MMM2025_Zero_shot},
or replacing the background with different ones
\cite{Chung_NeurIPS2022_HAT,Li_ICCV2023_StillMix,Wang_CVPR2021_RemovingBackground},
with customized architectures
\cite{Choi_NeurIPS2019_dancemall,Bae_ECCV2024_Devias,Duan_ECCV2022_Omnidebias}
to disentangle static appearance and dynamic action.
A different and promising approach is 
Appearance Free Dataset (AFD) \cite{Ilic_ECCV2022_RandomDot}. It first converts each frame of a video with random dots and warps the regions of moving people and objects using optical flow to completely remove the information of static appearance in the video and extract pure dynamic information of human action. Then they constructed a random dots video dataset AFD101 based on UCF101 \cite{Soomro_arXiv2012_UCF101}.
However, this approach has two crucial issues; computing flow and warping frames is computationally expensive, and models trained on the AFD101 dataset are not useful for RGB frames.

Instead, we propose to use edge frames as they contains less static appearance than RGB frames. 
Unlike AFD that needs an offline process for converting frames to random dots,
we extract edges on-the-fly during training
because edge detection is inexpensive to compute.
Before the advent of deep learning, edge detection-based action recognition was investigated \cite{Wang_MVA2016_arusingedge,Sappa_ICCS2006_EdgeMotionDetection,Suma_ISVC2008_SketchHumanAction}; however, no recent deep architectures for action recognition have explored this edge-based approach.

\subsection{Moment Exchange and Matching}

The proposed approach uses both RGB and edge frames as input for a two-stream model. The challenge is to handle the domain shift
between the RGB and edge frames.

Recent work on normalization
\cite{Ioffe_batch_normalization_ICLR2015,Ulyanov_arXiv2017_InstanceNormalization,Wu_group_normalization_ECCV2018,Huang_ICCV2017_AdaIN,Li_NeurIPS2019_PONO}
has shown that the
first and second moments (mean and variance) are simple but important information to reduce the domain shift.
Beyond normalization,
MoEx \cite{Li_CVPR2021_MoEx} computes and exchanges moments between two images
for data augmentation in the feature space, unlike image space augmentation \cite{Zhang_ICLR18_MixUp,Yun_arXiv2020_VideoMix,Kimata_MMAsia2022_ObjectMix}.

Inspired by the MoEx framework \cite{Li_CVPR2021_MoEx},
we perform domain adaptation by exchanging the moments of the features between layers in RGB and edge streams.

\section{Proposed Method}

Figure \ref{fig:architecture_all} shows the proposed model,
which has two streams of different ViT encoders,
one for RGB frames and the other for edge frames
computed with a lightweight online edge detector,
with a simple late fusion of frame-wise ViT features.
First, we perform different input normalization of RGB and edge frames, then apply domain adaptation by the proposed MoExDA module on intermediate features in the corresponding ViT layers between RGB and edge stream.

\subsection{Input Normalization}

Let $x_\mathrm{RGB} \in \R^{B \times T \times 3 \times H \times W}$ be the input minibatch of size $B$ of RGB video clips
with $T$ frames of size $H \times W$.
For the RGB stream, each frame is divided into $N$ patches to obtain the patch embedding $h^0_\mathrm{RGB} \in R^{B \times T \times (N+1) \times C}$, where $C$ is the embedding dimension.
For the edge stream, edge detection is performed on each frame of $x_\mathrm{RGB}$ to generate edge frames $x_\mathrm{edge} \in \mathbb{R}^{B \times T \times H \times W}$
and similarly the patch embeddings $h^0_\mathrm{edge} \in \mathbb{R}^{B \times T \times (N+1) \times C}$ are obtained.

A common practice is to scale the pixel values from $[0, 255]$ to $[0, 1]$ followed by normalization with the mean and standard deviation of the RGB channel;
$(\mu, \sigma) = ((0.485, 0.456, 0.406), (0.229, 0.224, 0.225))$.
These statistics are computed from images from ImageNet \cite{Deng_CVPR2009_ImageNet},
and are usually used for many tasks, including action recognition.
However, the distribution of pixel values in edge frames differs greatly from that of natural images in ImageNet or RGB frames of videos in action recognition datasets \cite{kay_arXiv2017_kinetics400,Soomro_arXiv2012_UCF101}, so the values are not appropriate.

To obtain statistics for edge frames,
we used all 265k frames of all 983 videos in the ``abseiling'' class of Kinetics400 dataset \cite{kay_arXiv2017_kinetics400} 
by grayscaling and edge detection with a Sobel filter.
The resulting mean and standard deviation of the pixel values of the
edge frames were
$(\mu, \sigma) =(0.026, 0.037)$.
We used these statistics for the edge stream in the experiments,
whereas the common statistics were used for the RGB stream.

\subsection{MoExDA}

Let the $l$ layer features in the
ViT encoder of the RGB and edge streams be
$h_\text{RGB}^{l}, h_\text{edge}^{l}
\in \mathbb{R}^{B \times T \times (N+1) \times C}$, respectively.
These features are computed from different domains;
$h_\text{RGB}^{l}$ are useful for classification
but are affected by static bias,
while $h_\text{edge}^{l}$ are considered
less affected by bias but thought to have low performance
due to information loss.

As shown on the right of Fig. \ref{fig:architecture_all},
the proposed MoExDA performs the interaction between the two streams
by aligning the first and second moments of the corresponding
features 
$h_\text{RGB}^{l}$ and $h_\text{edge}^{l}$,
which is inspired by MoEx \cite{Li_CVPR2021_MoEx}.

Let the means of
$h_\text{RGB}^{l}$ and $h_\text{edge}^{l}$ be
$\mu_\text{RGB}^{l}, \mu_\text{edge}^{l}$
and the standard deviations be
$\sigma_\text{RGB}^{l},  \sigma_\text{edge}^{l}$, respectively.
To align the moments 
of the edge features $h_\text{edge}^{l}$
to the RGB features $h_\text{RGB}^{l}$,
we exchange the moments as follows \cite{Li_CVPR2021_MoEx};
\begin{align}
    h_\text{edge}^{(\text{RGB})l}
    &=
    \frac{h_\text{edge}^{l} - \mu_\text{edge}^{l}}{\sigma_\text{edge}^{l}}
    \sigma_\text{RGB}^{l}
    + \mu_\text{RGB}^{l},
    \label{moex_Edge}
\end{align}
where $h_\text{edge}^{(\text{RGB})l}$ stands for the edge feature representation $h_\text{edge}^l$, while the moments are aligned to those of the RGB stream,
which is illustrated in Figure \ref{fig:architecture_all}.
This is referred to as the ``edge to RGB'' setting in the experiments,
as the \underline{edge} features are aligned \underline{to} the \underline{RGB} moments.

There are several design choices on how MoExDA is implemented,
as we discuss below.
The differences in performance according to each choice will be shown in the experiments.

\noindent\textbf{Normalization for moment computation.}
The first choice is in which dimensions the moments are calculated, that is, which normalization method is used,
and we investigated two methods.
The first is PONO (POsitional NOrmalization) \cite{Li_NeurIPS2019_PONO}, which calculates the moments in the channel dimension $C$ for each patch.
This is used in the original MoEx \cite{Li_CVPR2021_MoEx}. 
In this case,
the dimension of the moments of the RGB stream is
$\mu_\text{RGB}^{l}, \sigma_\text{RGB}^{l} \in \mathbb{R}^{B \times T \times (N+1)}$,
and the same applies to the edge stream.
The second is IN (Instance Normalization) \cite{Ulyanov_arXiv2017_InstanceNormalization}, which calculates moments in the spatial (or patch) dimension $N+1$,
therefore
$\mu_\text{RGB}^{l}, \sigma_\text{RGB}^{l} \in \mathbb{R}^{B \times T \times C}$.
The moment exchange Eq.\eqref{moex_Edge} is performed
with the tensor broadcast.

\noindent\textbf{Positions and layers.}
The second choice is where and how many MoExDA modules are inserted
into the Transformer block of the ViT encoder.
Fig. \ref{fig:architecture_all} shows
the setting we used for the experiments;
a single MoExDA module
inserted between the residual connection after MHA and the layer normalization of the MLP.

The number of layers in which the MoExDA modules are used
is also an important factor.
It will have a significant impact on performance
because
the distribution of the output of each layer will change
depending on the number of layers applied.
In our experiments, we compare settings where MoExDA modules are used
solely in the initial layer, or throughout all 12 layers.

\noindent\textbf{Directions of exchange.}
Eq.\eqref{moex_Edge} and Fig. \ref{fig:architecture_all}
show the moment exchange of the ``edge to RGB'' setting.
However, as the model has two streams, there are three possible directions for exchanging moments;
edge to RGB, RGB to edge, and bi-direction.
The first ``edge to RGB'' has been discussed before,
where the edge feature $h_\text{edge}^l$ is aligned to
the RGB moments.
In contrast, in ``RGB to edge'',
the RGB feature $h_\text{RGB}^l$ is aligned to the edge moments.

The third is a bidirectional moment exchange that performs these exchanges simultaneously.
In this case, we can expect to benefit from both directions of exchange.

\noindent\textbf{Stop gradient (sg).}
As information flows between two streams through
the MoExDA modules,
the gradient of one stream will be backpropagated to the other stream via the path of moment exchange.
This means that the gradients from the RGB stream might compromise the resistance to the static bias of the edge feature. In contrast, the RGB feature could be negatively impacted by the underperforming edge feature through the gradients.

In case of limiting the information exchange while maintaining the independence of each stream,
we stop gradients from propagating to the other stream via the moment exchange.

\section{Experimental results}

\noindent\textbf{Datasets.}
We constructed Kinetics50, a subset of Kinetics400 \cite{kay_arXiv2017_kinetics400} consisting of 50 classes of Mimetics \cite{Weinzaepfel_IJCV2021_Mimetics_dataset}
for efficiency of the experiments.
This consists of approximately 34k videos for training and 2.5k for validation.
During training, we used a video clip consisting of $T=16$ frames evenly sampled from each video, as input to the RGB stream.

\noindent\textbf{Loss Function.}
Since the proposed model has two streams,
we define the final loss as the weighted sum of the losses from each stream;
\begin{align}
    L_\text{{CE}}^\text{{all}} = \alpha_\text{{edge}} \ L_\text{{CE}}^\text{{edge}} + 
    \alpha_\text{{RGB}} \ L_\text{{CE}}^\text{{RGB}},
    \label{all_loss}
\end{align}
where 
$\alpha_\text{{RGB}} = 0.5$ and $\alpha_\text{{edge}} = 1.0$
are weights determined based on preliminary experiments.

\begin{table}[t]
    \centering
    \caption{
    Performance comparison for different
    normalizations, moment exchange directions, use of stop gradient, and streams.
    }
    \label{tab:performance_comparison}

    \resizebox{\linewidth}{!}{%
    \begin{tabular}{c@{\ }c@{\ }c@{\ }c@{\hspace{.5em}}c@{\hspace{.5em}}c@{\hspace{.5em}}c@{\hspace{.5em}}c}

    norm & ex dir & sg & stream & \shortstack{1\\ top-1$\uparrow$} & \shortstack{1$\sim$12\\ top-1$\uparrow$} & \shortstack{1$\sim$12\\BOR$\downarrow$} &
    \shortstack{1$\sim$12\\HOR$\uparrow$} \\
    \hline
    \multicolumn{3}{c}{baseline ViT} & edge & \multicolumn{2}{c}{68.84} & 73.43 & 23.83\\
    \multicolumn{3}{c}{baseline ViT} & RGB  & \multicolumn{2}{c}{82.65} & 83.66 & 29.39\\
    \multicolumn{3}{c}{S3Aug \cite{Sugiura_VISAPP2024_S3Aug}} & RGB  & \multicolumn{2}{c}{78.06} & 77.51 & 40.29\\
    \hline
    IN & Bidirection & \checkmark & edge & 71.22 & 33.74 & 42.63 & 11.63 \\
    IN & Bidirection &           & edge & 72.10 & 80.64 & 79.55 & 28.46 \\
    IN & RGB to edge & \checkmark & edge & 70.45 & 71.54 & 73.59 & 21.38 \\
    IN & RGB to edge &           & edge & 72.02 & 71.18 & 74.88 & 23.39 \\
    IN & edge to RGB & \checkmark & edge & 73.03 & 81.80 & 80.47 & 30.43 \\
    IN & edge to RGB &           & edge & 72.70 & 82.77 & 79.51 & 31.40 \\
    IN & Bidirection & \checkmark & RGB  & 77.94 & 32.45 & 56.88 & 18.64 \\
    IN & Bidirection &           & RGB  & 78.34 & 80.68 & 80.11 & 27.62 \\
    IN & RGB to edge & \checkmark & RGB  & 78.86 & 75.00 & 78.02 & 22.06 \\
    IN & RGB to edge &           & RGB  & 79.75 & 73.07 & 78.58 & 24.44 \\
    IN & edge to RGB & \checkmark & RGB  & 82.21 & 82.29 & 80.92 & 30.11 \\
    IN & edge to RGB &           & RGB  & 82.37 & 83.05 & 79.31 & 31.44 \\
    PONO & Bidirection & \checkmark & edge & 74.36 & 73.11 & 74.11 & 25.36 \\
    PONO & Bidirection &           & edge & 74.52 & 74.44 & 73.75 & 23.63 \\
    PONO & RGB to edge & \checkmark & edge & 71.94 & 71.78 & 69.48 & 22.54 \\
    PONO & RGB to edge &           & edge & 71.22 & 71.86 & 72.87 & 22.63 \\
    PONO & edge to RGB & \checkmark & edge & 72.91 & 75.12 & 71.09 & 23.39 \\
    PONO & edge to RGB &           & edge & 74.36 & 75.89 & 75.20 & 26.53 \\
    PONO & Bidirection & \checkmark & RGB  & 79.34 & 79.63 & 80.07 & 27.54 \\
    PONO & Bidirection &           & RGB  & 79.15 & 79.87 & 79.23 & 28.42 \\
    PONO & RGB to edge & \checkmark & RGB  & 79.43 & 79.03 & 79.23 & 27.25 \\
    PONO & RGB to edge &           & RGB  & 79.87 & 80.31 & 80.03 & 26.61 \\
    PONO & edge to RGB & \checkmark & RGB  & 82.33 & 81.64 & 80.27 & 28.58 \\
    PONO & edge to RGB &           & RGB  & 81.72 & 81.64 & 81.12 & 30.07 \\
    \hline
    \end{tabular}%
    }
\end{table}

\begin{figure}[t]
    \centering

    \includegraphics[width=\linewidth]{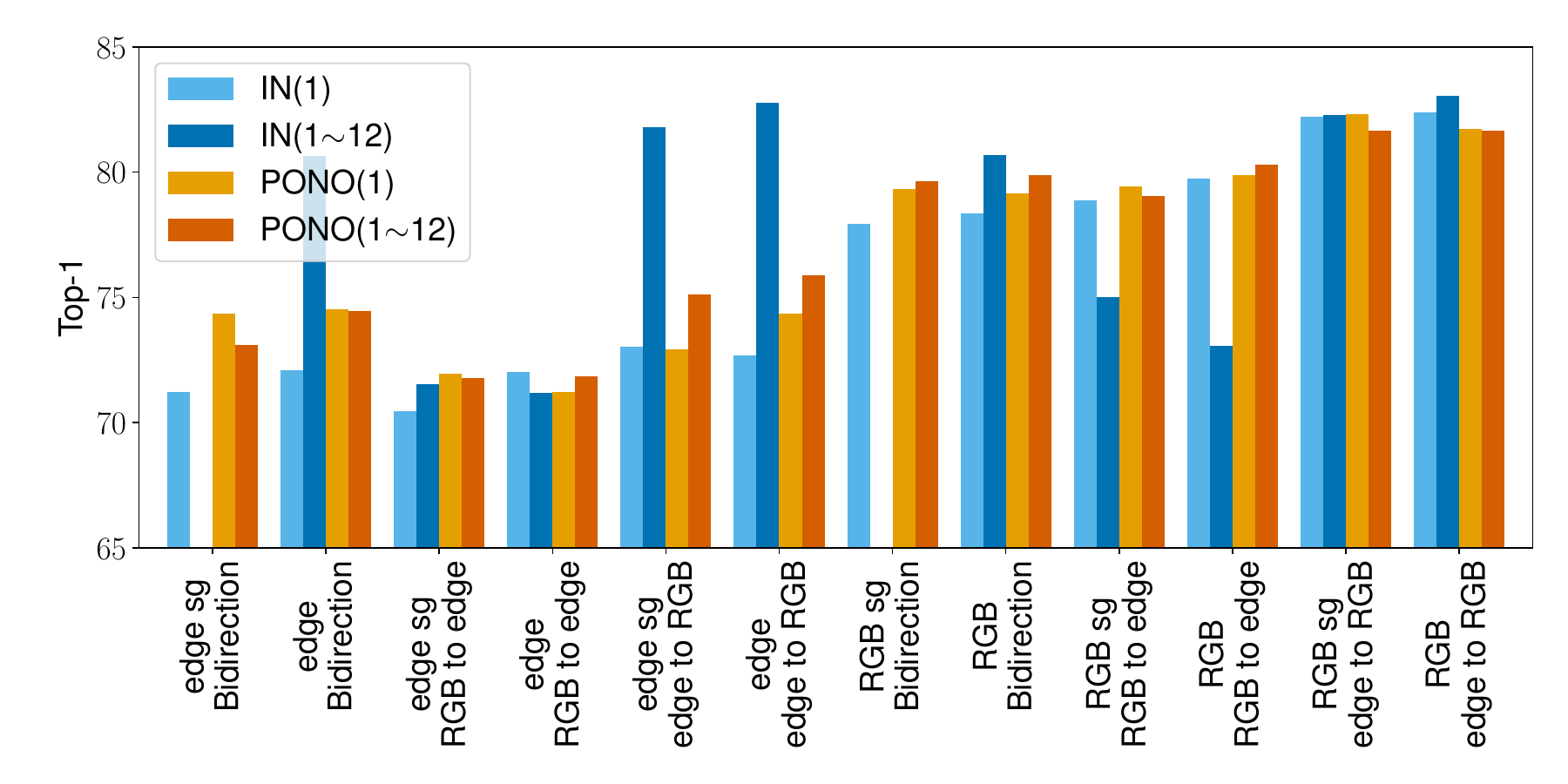}%

    \includegraphics[width=\linewidth]{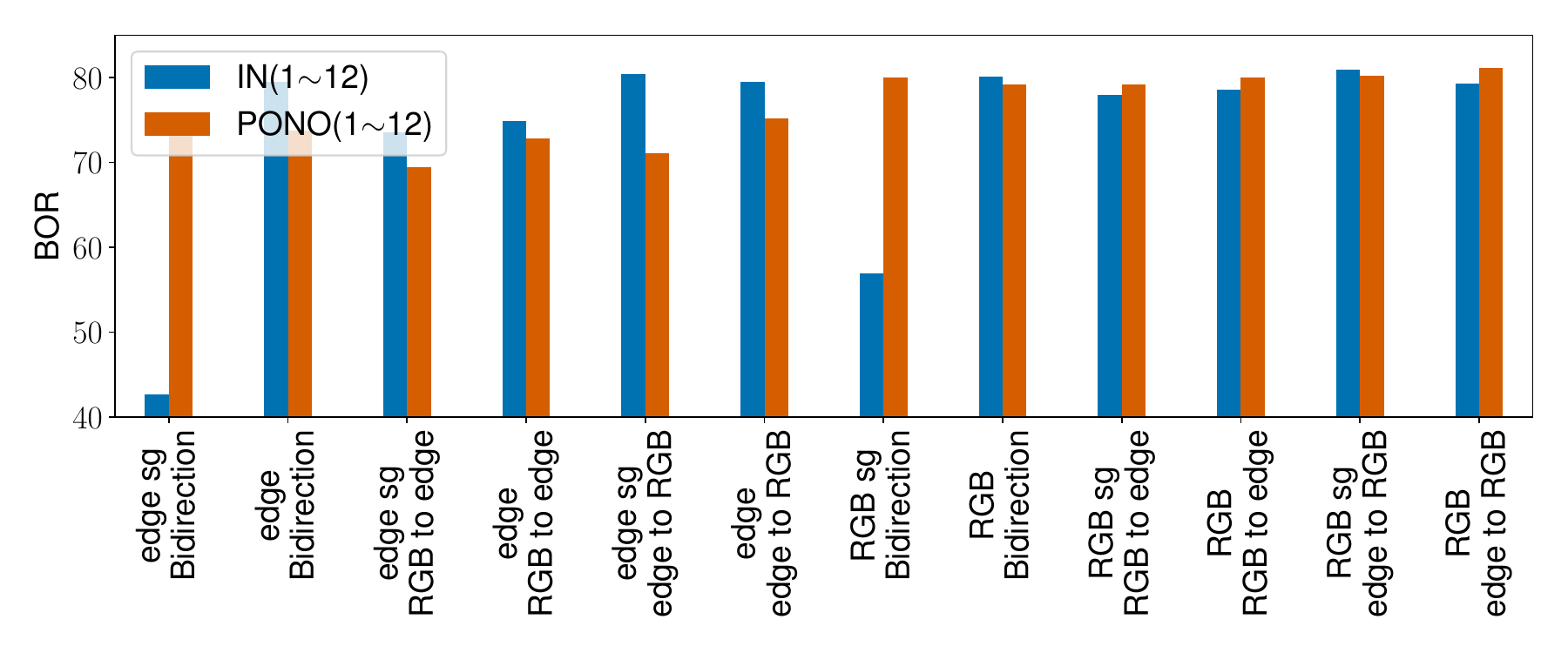}%

    \includegraphics[width=\linewidth]{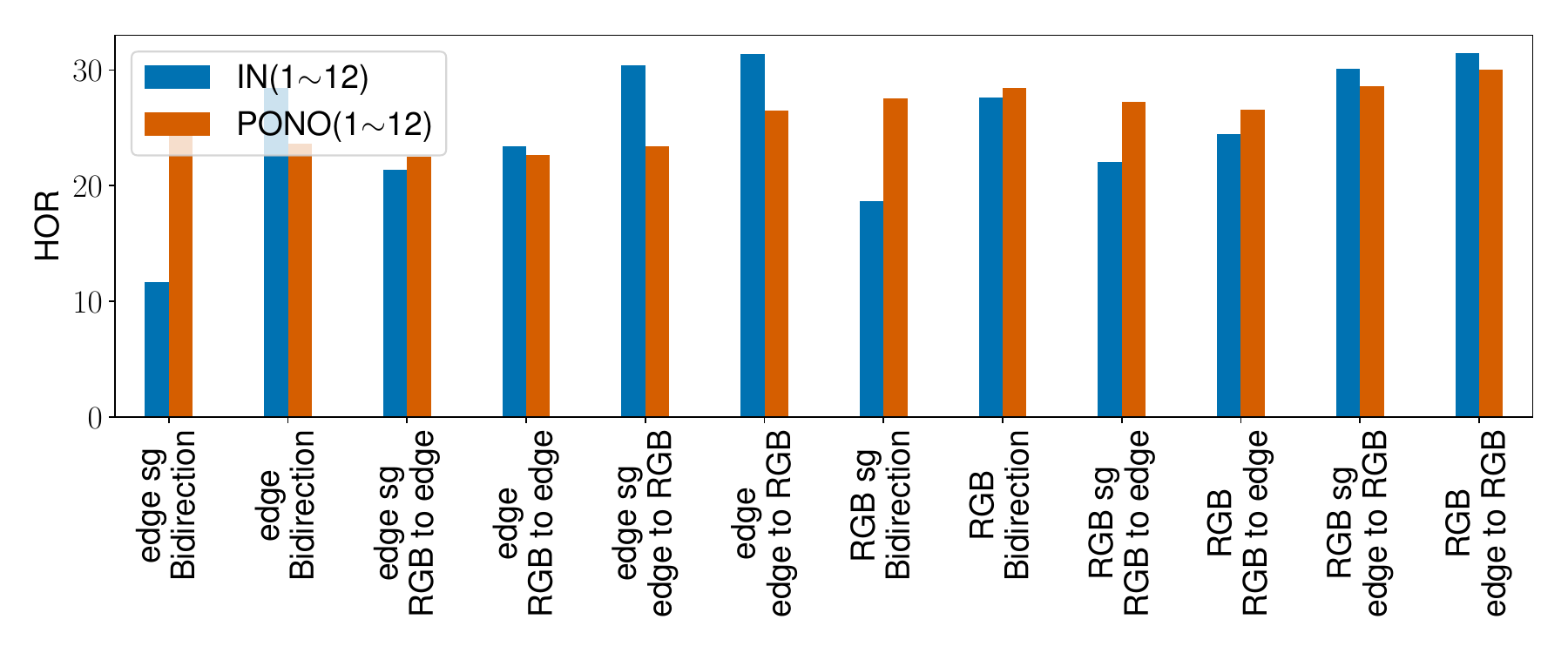}%
    
    \caption{
        Comparisons of 
        (top) top-1 performance,
        (middle) BOR, and
        (bottom) HOR for different configurations.
    }
    \label{fig:top1_vs_configurations}
    
\end{figure}

\noindent\textbf{Performance comparison.}
We report the top-1 performances of both RGB and edge streams.
Table \ref{tab:performance_comparison} shows the comparison
of different configurations.
The first three lines are baseline ViTs with the same late fusion and prior work \cite{Sugiura_VISAPP2024_S3Aug},
with a single stream that takes RGB or edge frames as input.

In our performance evaluation, the column ``1 top-1'' indicates performance when MoExDA is inserted into the first layer only, while the ``1$\sim$12 top-1'' column shows results when MoExDA is inserted across all layers. Using either IN or PONO normalization, the RGB stream consistently outperforms the edge stream. Figure \ref{fig:top1_vs_configurations} clearly illustrates this pattern, with the RGB stream performance (right side) exceeding the edge stream performance (left side). This performance difference is reasonable, as the edge stream's input contains minimal static bias. However, in all cases, our proposed method achieves an improvement in top-1 performance by 2\% to 5\% over the baseline ViT that uses only edge input, confirming the effectiveness of MoExDA's information exchange from RGB to edge stream.

When comparing the two normalization methods, IN and PONO show no significant performance differences, although PONO shows slight advantages in most cases. As Figure 3 demonstrates, IN is considerably more sensitive to configuration changes than PONO, making PONO the more practical choice for various analyses and hyperparameter searches. Note that IN shows severe performance degradation under specific conditions (edge stream, bi-direction, with sg), and analysis of this phenomenon remains a future research topic.

Furthermore, IN and PONO exhibit distinct performance differences based on the direction of moment exchange. IN tends to perform better with the RGB-to-edge direction, while PONO often shows better results with edge-to-RGB direction. This difference comes from the way each normalization calculates the moments. PONO normalizes locally at each image position, whereas IN performs global normalization over the entire image. Though this difference has little effect on standard RGB images, when processing edges, which contain concentrated local information, PONO's edge-to-RGB approach becomes particularly effective by matching local edge moments with RGB local moments.
Note that the presence or absence of stop gradients did not show a significant impact on performance.

\noindent\textbf{Evaluation of static bias.}
We evaluated the effectiveness of the proposed method using BOR (Background-Only Ratio) and HOR (Human-Only Ratio), which are evaluation metrics for static bias \cite{Chung_NeurIPS2022_HAT}, compared with existing approaches.
The result are shown in Tab. \ref{tab:performance_comparison} 
and Fig. \ref{fig:top1_vs_configurations}, where MoExDA is inserted across all layers (excluding baseline ViT and S3Aug). The ViT baseline with the RGB stream achieves the highest top-1 performance but shows a high BOR, indicating that it is strongly influenced by background bias. Similarly, while the conventional S3Aug method achieves the highest HOR, it suffers from high BOR measurements.

The proposed method typically achieves lower BOR values. Without special cases of IN performance degradation, the lowest BOR falls below 70\%, which is significantly lower than the baseline. Although the method achieved an HOR of 30\%, exceeding the baseline ViT, this remains below the HOR value of the conventional S3Aug.  Resolving this trade-off will be a future research topic. Overall, while the proposed method effectively reduces background bias, further improvements are needed in action prediction that focuses on human actors.

\section{Conclusion}

In this paper, we proposed an edge-based action recognition approach that suppresses static bias effects and the MoExDA module to bridge the domain gap between RGB and edge streams. Although the experimental results require careful interpretation, our method achieves a superior BOR compared to the baselines. Future work includes examining the effectiveness of applying MoExDA across different layers and developing techniques to estimate moments during training (similar to BN \cite{Ioffe_batch_normalization_ICLR2015}), allowing the use of only the edge stream during evaluation.

\noindent\textbf{Acknowledgments.}
This work was supported in part by
JSPS KAKENHI Grant Numbers P22K12090 and JP25K03138.

{
    \bibliographystyle{IEEEtran}
    \bibliography{mybib,all}

\begin{thebibliography}{10}
\providecommand{\url}[1]{#1}
\csname url@samestyle\endcsname
\providecommand{\newblock}{\relax}
\providecommand{\bibinfo}[2]{#2}
\providecommand{\BIBentrySTDinterwordspacing}{\spaceskip=0pt\relax}
\providecommand{\BIBentryALTinterwordstretchfactor}{4}
\providecommand{\BIBentryALTinterwordspacing}{\spaceskip=\fontdimen2\font plus
\BIBentryALTinterwordstretchfactor\fontdimen3\font minus \fontdimen4\font\relax}
\providecommand{\BIBforeignlanguage}[2]{{%
\expandafter\ifx\csname l@#1\endcsname\relax
\typeout{** WARNING: IEEEtran.bst: No hyphenation pattern has been}%
\typeout{** loaded for the language `#1'. Using the pattern for}%
\typeout{** the default language instead.}%
\else
\language=\csname l@#1\endcsname
\fi
#2}}
\providecommand{\BIBdecl}{\relax}
\BIBdecl

\bibitem{Yu_2022_human_action_recog_survey}
\BIBentryALTinterwordspacing
Y.~Kong and Y.~Fu, ``Human action recognition and prediction: A survey,'' \emph{International Journal of Computer Vision}, vol. 130, no.~5, pp. 1366--1401, may 2022. [Online]. Available: \url{https://doi.org/10.1007/s11263-022-01594-9}
\BIBentrySTDinterwordspacing

\bibitem{Kong_IJCV2022_actionrecognitionsurvey}
------, ``Human action recognition and prediction: A survey,'' \emph{International Journal of Computer Vision}, vol. 130, no.~5, pp. 1366--1401, 2022.

\bibitem{Hutchinson_IEEEAccess2021_Action_Recognition_Survey}
M.~S. Hutchinson and V.~N. Gadepally, ``Video action understanding,'' \emph{IEEE Access}, vol.~9, pp. 134\,611--134\,637, 2021.

\bibitem{Hara_2018CVPR_3D_ResNet}
K.~Hara, H.~Kataoka, and Y.~Satoh, ``Can spatiotemporal 3d cnns retrace the history of 2d cnns and imagenet?'' in \emph{Proceedings of the IEEE Conference on Computer Vision and Pattern Recognition (CVPR)}, June 2018.

\bibitem{Chung_NeurIPS2022_HAT}
\BIBentryALTinterwordspacing
J.~Chung, Y.~Wu, and O.~Russakovsky, ``Enabling detailed action recognition evaluation through video dataset augmentation,'' in \emph{Thirty-sixth Conference on Neural Information Processing Systems Datasets and Benchmarks Track}, 2022. [Online]. Available: \url{https://openreview.net/forum?id=eOnQ2etkxto}
\BIBentrySTDinterwordspacing

\bibitem{Li_ECCV2018_RESOUND}
Y.~Li, Y.~Li, and N.~Vasconcelos, ``Resound: Towards action recognition without representation bias,'' in \emph{Proceedings of the European Conference on Computer Vision (ECCV)}, September 2018.

\bibitem{Fukuzawa_MMM2025_Zero_shot}
T.~Fukuzawa, K.~Hara, H.~Kataoka, and T.~Tamaki, ``Can masking background and object reduce static bias for zero-shot action recognition?'' in \emph{MultiMedia Modeling}, I.~Ide, I.~Kompatsiaris, C.~Xu, K.~Yanai, W.-T. Chu, N.~Nitta, M.~Riegler, and T.~Yamasaki, Eds.\hskip 1em plus 0.5em minus 0.4em\relax Singapore: Springer Nature Singapore, 2025, pp. 366--379.

\bibitem{He_ECCV2016_Workshops_withouthuman}
Y.~He, S.~Shirakabe, Y.~Satoh, and H.~Kataoka, ``Human action recognition without human,'' in \emph{Computer Vision -- ECCV 2016 Workshops}, G.~Hua and H.~J{\'e}gou, Eds.\hskip 1em plus 0.5em minus 0.4em\relax Cham: Springer International Publishing, 2016, pp. 11--17.

\bibitem{Jain_CVPR2015_WhatdoObject}
M.~Jain, J.~C. van Gemert, and C.~G.~M. Snoek, ``What do 15,000 object categories tell us about classifying and localizing actions?'' in \emph{Proceedings of the IEEE Conference on Computer Vision and Pattern Recognition (CVPR)}, June 2015.

\bibitem{Wu_CVPR2016_HarnessingObject}
Z.~Wu, Y.~Fu, Y.-G. Jiang, and L.~Sigal, ``Harnessing object and scene semantics for large-scale video understanding,'' in \emph{Proceedings of the IEEE Conference on Computer Vision and Pattern Recognition (CVPR)}, June 2016.

\bibitem{Ilic_ECCV2022_RandomDot}
F.~Ilic, T.~Pock, and R.~P. Wildes, ``Is appearance free action recognition possible?'' in \emph{Computer Vision -- ECCV 2022}, S.~Avidan, G.~Brostow, M.~Ciss{\'e}, G.~M. Farinella, and T.~Hassner, Eds.\hskip 1em plus 0.5em minus 0.4em\relax Cham: Springer Nature Switzerland, 2022, pp. 156--173.

\bibitem{Dosovitskiy_ICLR2021_ViT_Vision_transformer}
\BIBentryALTinterwordspacing
A.~Dosovitskiy, L.~Beyer, A.~Kolesnikov, D.~Weissenborn, X.~Zhai, T.~Unterthiner, M.~Dehghani, M.~Minderer, G.~Heigold, S.~Gelly, J.~Uszkoreit, and N.~Houlsby, ``An image is worth 16x16 words: Transformers for image recognition at scale,'' in \emph{International Conference on Learning Representations}, 2021. [Online]. Available: \url{https://openreview.net/forum?id=YicbFdNTTy}
\BIBentrySTDinterwordspacing

\bibitem{Li_CVPR2021_MoEx}
B.~Li, F.~Wu, S.-N. Lim, S.~Belongie, and K.~Q. Weinberger, ``On feature normalization and data augmentation,'' in \emph{Proceedings of the IEEE/CVF Conference on Computer Vision and Pattern Recognition (CVPR)}, June 2021, pp. 12\,383--12\,392.

\bibitem{Carreira_2017CVPR_I3D}
J.~Carreira and A.~Zisserman, ``Quo vadis, action recognition? a new model and the kinetics dataset,'' in \emph{Proceedings of the IEEE Conference on Computer Vision and Pattern Recognition (CVPR)}, July 2017.

\bibitem{Feichtenhofer_2020CVPR_X3D}
\BIBentryALTinterwordspacing
C.~Feichtenhofer, ``X3d: Expanding architectures for efficient video recognition,'' in \emph{Proceedings of the IEEE/CVF Conference on Computer Vision and Pattern Recognition (CVPR)}, June 2020. [Online]. Available: \url{https://openaccess.thecvf.com/content_CVPR_2020/html/Feichtenhofer_X3D_Expanding_Architectures_for_Efficient_Video_Recognition_CVPR_2020_paper.html}
\BIBentrySTDinterwordspacing

\bibitem{Feichtenhofer_2019ICCV_SlowFast}
C.~Feichtenhofer, H.~Fan, J.~Malik, and K.~He, ``Slowfast networks for video recognition,'' in \emph{Proceedings of the IEEE/CVF International Conference on Computer Vision (ICCV)}, October 2019.

\bibitem{Hara_IEICE_ED2020_Action_Recognition_Survey}
K.~Hara, ``Recent advances in video action recognition with 3d convolutions,'' \emph{IEICE Transactions on Fundamentals of Electronics, Communications and Computer Sciences}, vol. E104.A, no.~6, pp. 846--856, 2021.

\bibitem{Arnab_2021_ICCV_ViVit}
A.~Arnab, M.~Dehghani, G.~Heigold, C.~Sun, M.~Lu\v{c}i\'c, and C.~Schmid, ``Vivit: A video vision transformer,'' in \emph{Proceedings of the IEEE/CVF International Conference on Computer Vision (ICCV)}, October 2021, pp. 6836--6846.

\bibitem{Bertasius_ICML2021_TimeSformer}
\BIBentryALTinterwordspacing
G.~Bertasius, H.~Wang, and L.~Torresani, ``Is space-time attention all you need for video understanding?'' in \emph{Proceedings of the 38th International Conference on Machine Learning}, ser. Proceedings of Machine Learning Research, M.~Meila and T.~Zhang, Eds., vol. 139.\hskip 1em plus 0.5em minus 0.4em\relax PMLR, 18--24 Jul 2021, pp. 813--824. [Online]. Available: \url{https://proceedings.mlr.press/v139/bertasius21a.html}
\BIBentrySTDinterwordspacing

\bibitem{Selva_TPAMI2023_VideoTransformerSurvey}
J.~Selva, A.~S. Johansen, S.~Escalera, K.~Nasrollahi, T.~B. Moeslund, and A.~Clapes, ``Video transformers: A survey,'' \emph{IEEE Transactions on Pattern Analysis \&amp; Machine Intelligence}, vol.~45, no.~11, pp. 12\,922--12\,943, nov 2023.

\bibitem{Liu_2022CVPR_VideoSwin}
Z.~Liu, J.~Ning, Y.~Cao, Y.~Wei, Z.~Zhang, S.~Lin, and H.~Hu, ``Video swin transformer,'' in \emph{Proceedings of the IEEE/CVF Conference on Computer Vision and Pattern Recognition (CVPR)}, June 2022, pp. 3202--3211.

\bibitem{Xu-EMNLP2021-VideoCLIP}
\BIBentryALTinterwordspacing
H.~Xu, G.~Ghosh, P.-Y. Huang, D.~Okhonko, A.~Aghajanyan, F.~Metze, L.~Zettlemoyer, and C.~Feichtenhofer, ``{V}ideo{CLIP}: Contrastive pre-training for zero-shot video-text understanding,'' in \emph{Proceedings of the 2021 Conference on Empirical Methods in Natural Language Processing}.\hskip 1em plus 0.5em minus 0.4em\relax Online and Punta Cana, Dominican Republic: Association for Computational Linguistics, Nov. 2021, pp. 6787--6800. [Online]. Available: \url{https://aclanthology.org/2021.emnlp-main.544}
\BIBentrySTDinterwordspacing

\bibitem{Wang_arxiv2021_ActionCLIP}
\BIBentryALTinterwordspacing
M.~Wang, J.~Xing, and Y.~Liu, ``Actionclip: {A} new paradigm for video action recognition,'' \emph{CoRR}, vol. abs/2109.08472, 2021. [Online]. Available: \url{https://arxiv.org/abs/2109.08472}
\BIBentrySTDinterwordspacing

\bibitem{Rasheed_CVPR2023_Vifi-CLIP}
H.~Rasheed, M.~U. Khattak, M.~Maaz, S.~Khan, and F.~S. Khan, ``Fine-tuned clip models are efficient video learners,'' in \emph{Proceedings of the IEEE/CVF Conference on Computer Vision and Pattern Recognition (CVPR)}, June 2023, pp. 6545--6554.

\bibitem{Sugiura_VISAPP2024_S3Aug}
\BIBentryALTinterwordspacing
T.~Sugiura and T.~Tamaki, ``S3aug: Segmentation, sampling, and shift for action recognition,'' in \emph{Proceedings of the 19th International Joint Conference on Computer Vision, Imaging and Computer Graphics Theory and Applications - Volume 2: VISAPP}, INSTICC.\hskip 1em plus 0.5em minus 0.4em\relax SciTePress, 2024, pp. 71--79. [Online]. Available: \url{https://www.scitepress.org/Link.aspx?doi=10.5220/0012310400003660}
\BIBentrySTDinterwordspacing

\bibitem{Li_ICCV2023_StillMix}
H.~Li, Y.~Liu, H.~Zhang, and B.~Li, ``Mitigating and evaluating static bias of action representations in the background and the foreground,'' in \emph{Proceedings of the IEEE/CVF International Conference on Computer Vision (ICCV)}, October 2023, pp. 19\,911--19\,923.

\bibitem{Wang_CVPR2021_RemovingBackground}
J.~Wang, Y.~Gao, K.~Li, Y.~Lin, A.~J. Ma, H.~Cheng, P.~Peng, F.~Huang, R.~Ji, and X.~Sun, ``Removing the background by adding the background: Towards background robust self-supervised video representation learning,'' in \emph{Proceedings of the IEEE/CVF Conference on Computer Vision and Pattern Recognition (CVPR)}, June 2021, pp. 11\,804--11\,813.

\bibitem{Choi_NeurIPS2019_dancemall}
\BIBentryALTinterwordspacing
J.~Choi, C.~Gao, J.~C.~E. Messou, and J.-B. Huang, ``Why can't {I} dance in the mall? learning to mitigate scene bias in action recognition,'' in \emph{Advances in Neural Information Processing Systems}, H.~Wallach, H.~Larochelle, A.~Beygelzimer, F.~d\textquotesingle Alch\'{e}-Buc, E.~Fox, and R.~Garnett, Eds., vol.~32.\hskip 1em plus 0.5em minus 0.4em\relax Curran Associates, Inc., 2019. [Online]. Available: \url{https://proceedings.neurips.cc/paper_files/paper/2019/file/ab817c9349cf9c4f6877e1894a1faa00-Paper.pdf}
\BIBentrySTDinterwordspacing

\bibitem{Bae_ECCV2024_Devias}
K.~Bae, G.~Ahn, Y.~Kim, and J.~Choi, ``Devias: Learning disentangled video representations of action and scene,'' in \emph{Computer Vision -- ECCV 2024}, A.~Leonardis, E.~Ricci, S.~Roth, O.~Russakovsky, T.~Sattler, and G.~Varol, Eds.\hskip 1em plus 0.5em minus 0.4em\relax Cham: Springer Nature Switzerland, 2025, pp. 431--448.

\bibitem{Duan_ECCV2022_Omnidebias}
H.~Duan, Y.~Zhao, K.~Chen, Y.~Xiong, and D.~Lin, ``Mitigating representation bias in action recognition: Algorithms and benchmarks,'' in \emph{Computer Vision -- ECCV 2022 Workshops}, L.~Karlinsky, T.~Michaeli, and K.~Nishino, Eds.\hskip 1em plus 0.5em minus 0.4em\relax Cham: Springer Nature Switzerland, 2023, pp. 557--575.

\bibitem{Soomro_arXiv2012_UCF101}
\BIBentryALTinterwordspacing
K.~Soomro, A.~R. Zamir, and M.~Shah, ``{UCF101:} {A} dataset of 101 human actions classes from videos in the wild,'' \emph{CoRR}, vol. abs/1212.0402, 2012. [Online]. Available: \url{http://arxiv.org/abs/1212.0402}
\BIBentrySTDinterwordspacing

\bibitem{Wang_MVA2016_arusingedge}
X.~Wang and C.~Qi, ``Action recognition using edge trajectories and motion acceleration descriptor,'' \emph{Machine Vision and Applications}, vol.~27, no.~6, pp. 861--875, 2016.

\bibitem{Sappa_ICCS2006_EdgeMotionDetection}
A.~D. Sappa and F.~Dornaika, ``An edge-based approach to motion detection,'' in \emph{Computational Science -- ICCS 2006}, V.~N. Alexandrov, G.~D. van Albada, P.~M.~A. Sloot, and J.~Dongarra, Eds.\hskip 1em plus 0.5em minus 0.4em\relax Berlin, Heidelberg: Springer Berlin Heidelberg, 2006, pp. 563--570.

\bibitem{Suma_ISVC2008_SketchHumanAction}
E.~A. Suma, C.~W. Sinclair, J.~Babbs, and R.~Souvenir, ``A sketch-based approach for detecting common human actions,'' in \emph{Advances in Visual Computing}, G.~Bebis, R.~Boyle, B.~Parvin, D.~Koracin, P.~Remagnino, F.~Porikli, J.~Peters, J.~Klosowski, L.~Arns, Y.~K. Chun, T.-M. Rhyne, and L.~Monroe, Eds.\hskip 1em plus 0.5em minus 0.4em\relax Berlin, Heidelberg: Springer Berlin Heidelberg, 2008, pp. 418--427.

\bibitem{Ioffe_batch_normalization_ICLR2015}
\BIBentryALTinterwordspacing
S.~Ioffe and C.~Szegedy, ``\BIBforeignlanguage{en}{Batch {Normalization}: {Accelerating} {Deep} {Network} {Training} by {Reducing} {Internal} {Covariate} {Shift}},'' in \emph{\BIBforeignlanguage{en}{Proceedings of the 32nd {International} {Conference} on {Machine} {Learning}}}.\hskip 1em plus 0.5em minus 0.4em\relax PMLR, Jun. 2015, pp. 448--456, batch Normalization BN. [Online]. Available: \url{https://proceedings.mlr.press/v37/ioffe15.html}
\BIBentrySTDinterwordspacing

\bibitem{Ulyanov_arXiv2017_InstanceNormalization}
\BIBentryALTinterwordspacing
D.~Ulyanov, A.~Vedaldi, and V.~Lempitsky, ``Instance normalization: The missing ingredient for fast stylization,'' 2017. [Online]. Available: \url{https://arxiv.org/abs/1607.08022}
\BIBentrySTDinterwordspacing

\bibitem{Wu_group_normalization_ECCV2018}
\BIBentryALTinterwordspacing
Y.~Wu and K.~He, ``\BIBforeignlanguage{en}{Group {Normalization}},'' in \emph{\BIBforeignlanguage{en}{Computer {Vision} ^^e2^^80^^93 {ECCV} 2018}}, V.~Ferrari, M.~Hebert, C.~Sminchisescu, and Y.~Weiss, Eds.\hskip 1em plus 0.5em minus 0.4em\relax Cham: Springer International Publishing, 2018, pp. 3--19. [Online]. Available: \url{https://link.springer.com/chapter/10.1007/978-3-030-01261-8_1}
\BIBentrySTDinterwordspacing

\bibitem{Huang_ICCV2017_AdaIN}
X.~Huang and S.~Belongie, ``Arbitrary style transfer in real-time with adaptive instance normalization,'' in \emph{Proceedings of the IEEE International Conference on Computer Vision (ICCV)}, Oct 2017.

\bibitem{Li_NeurIPS2019_PONO}
\BIBentryALTinterwordspacing
B.~Li, F.~Wu, K.~Q. Weinberger, and S.~Belongie, ``Positional normalization,'' in \emph{Advances in Neural Information Processing Systems}, H.~Wallach, H.~Larochelle, A.~Beygelzimer, F.~d\textquotesingle Alch\'{e}-Buc, E.~Fox, and R.~Garnett, Eds., vol.~32.\hskip 1em plus 0.5em minus 0.4em\relax Curran Associates, Inc., 2019. [Online]. Available: \url{https://proceedings.neurips.cc/paper_files/paper/2019/file/6d0f846348a856321729a2f36734d1a7-Paper.pdf}
\BIBentrySTDinterwordspacing

\bibitem{Zhang_ICLR18_MixUp}
\BIBentryALTinterwordspacing
H.~Zhang, M.~Ciss{\'{e}}, Y.~N. Dauphin, and D.~Lopez{-}Paz, ``mixup: Beyond empirical risk minimization,'' in \emph{6th International Conference on Learning Representations, {ICLR} 2018, Vancouver, BC, Canada, April 30 - May 3, 2018, Conference Track Proceedings}.\hskip 1em plus 0.5em minus 0.4em\relax OpenReview.net, 2018. [Online]. Available: \url{https://openreview.net/forum?id=r1Ddp1-Rb}
\BIBentrySTDinterwordspacing

\bibitem{Yun_arXiv2020_VideoMix}
\BIBentryALTinterwordspacing
S.~Yun, S.~J. Oh, B.~Heo, D.~Han, and J.~Kim, ``Videomix: Rethinking data augmentation for video classification,'' 2020. [Online]. Available: \url{https://arxiv.org/abs/2012.03457}
\BIBentrySTDinterwordspacing

\bibitem{Kimata_MMAsia2022_ObjectMix}
\BIBentryALTinterwordspacing
J.~Kimata, T.~Nitta, and T.~Tamaki, ``Objectmix: Data augmentation by copy-pasting objects in videos for action recognition,'' in \emph{Proceedings of the 4th ACM International Conference on Multimedia in Asia}, ser. MMAsia '22.\hskip 1em plus 0.5em minus 0.4em\relax New York, NY, USA: Association for Computing Machinery, 2022. [Online]. Available: \url{https://doi.org/10.1145/3551626.3564941}
\BIBentrySTDinterwordspacing

\bibitem{Deng_CVPR2009_ImageNet}
J.~Deng, W.~Dong, R.~Socher, L.-J. Li, K.~Li, and L.~Fei-Fei, ``Imagenet: A large-scale hierarchical image database,'' in \emph{2009 IEEE Conference on Computer Vision and Pattern Recognition}, 2009, pp. 248--255.

\bibitem{kay_arXiv2017_kinetics400}
\BIBentryALTinterwordspacing
W.~Kay, J.~Carreira, K.~Simonyan, B.~Zhang, C.~Hillier, S.~Vijayanarasimhan, F.~Viola, T.~Green, T.~Back, P.~Natsev, M.~Suleyman, and A.~Zisserman, ``The kinetics human action video dataset,'' \emph{CoRR}, vol. abs/1705.06950, 2017. [Online]. Available: \url{http://arxiv.org/abs/1705.06950}
\BIBentrySTDinterwordspacing

\bibitem{Weinzaepfel_IJCV2021_Mimetics_dataset}
\BIBentryALTinterwordspacing
P.~Weinzaepfel and G.~Rogez, ``Mimetics: Towards understanding human actions out of context,'' \emph{International Journal of Computer Vision}, vol. 129, no.~5, pp. 1675--1690, 2021. [Online]. Available: \url{https://doi.org/10.1007/s11263-021-01446-y}
\BIBentrySTDinterwordspacing

\end{thebibliography}
}

\end{document}